\documentclass[conference]{IEEEtran}
\IEEEoverridecommandlockouts
\usepackage{cite}
\usepackage{amsmath,amssymb,amsfonts}
\usepackage{algorithmic}
\usepackage{graphicx}
\usepackage{textcomp}
\usepackage{xcolor}

\usepackage{url}
\usepackage{multirow}

\def\BibTeX{{\rm B\kern-.05em{\sc i\kern-.025em b}\kern-.08em
    T\kern-.1667em\lower.7ex\hbox{E}\kern-.125emX}}
    
\begin{document}

\title{Optimising Resource Management for \\ Embedded Machine Learning\\
}

\author{\IEEEauthorblockN{Lei Xun, Long Tran-Thanh, Bashir M Al-Hashimi, Geoff V. Merrett}
\IEEEauthorblockA{\textit{School of Electronics and Computer Science} \\
\textit{University of Southampton}\\
Southampton, UK \\
{\{lx2u16, ltt08r, bmah, gvm\}}@ecs.soton.ac.uk}

}

\maketitle

\begin{abstract}
Machine learning inference is increasingly being executed locally on mobile and embedded platforms, due to the clear advantages in latency, privacy and connectivity. In this paper, we present approaches for online resource management in heterogeneous multi-core systems and show how they can be applied to optimise the performance of machine learning workloads. Performance can be defined using platform-dependent (e.g. speed, energy) and platform-independent (accuracy, confidence) metrics. In particular, we show how a Deep Neural Network (DNN) can be dynamically scalable to trade-off these various performance metrics. Achieving consistent performance when executing on different platforms is necessary yet challenging, due to the different resources provided and their capability, and their time-varying availability when executing alongside other workloads. Managing the interface between available hardware resources (often numerous and heterogeneous in nature), software requirements, and user experience is increasingly complex.\\
\end{abstract}


\section{Introduction}
Deep Neural Networks (DNNs)\cite{deeplearning} are widely used in many applications including computer vision \cite{alexnet,yolo} and natural language processing\cite{vaswani2017attention}. Compared to traditional hand-engineered machine learning algorithms, DNNs have demonstrated near or super-human accuracy.

\begin{figure}
\centerline{\includegraphics[width=\columnwidth]{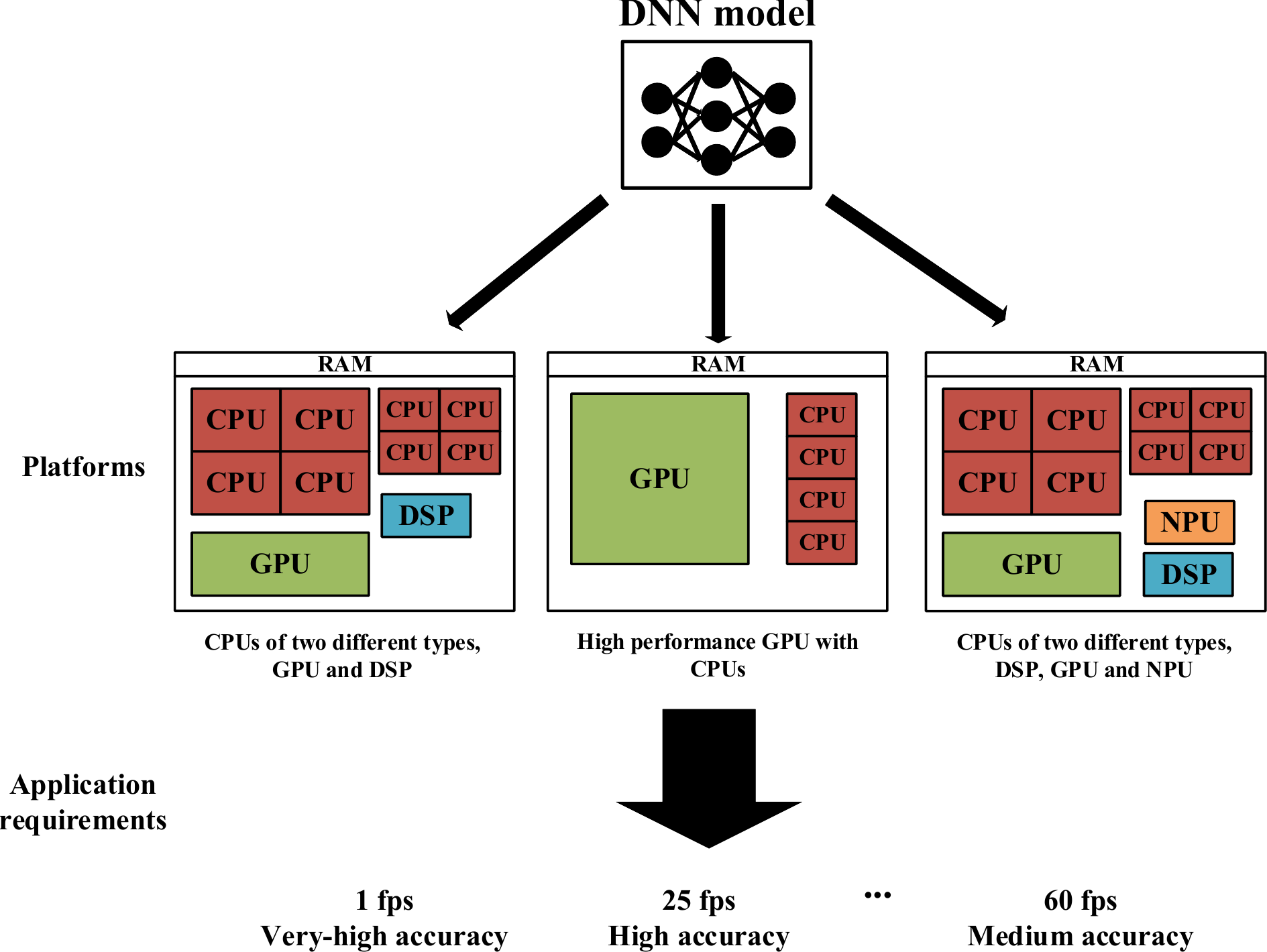}}
\caption{DNNs can be deployed on a variety of hardware platforms with different computing resources. At design time, the DNN is compressed using techniques such as static model pruning \cite{NetAdapt, he2018amc} (see Section \ref{review}), and then mapped onto different computing resources to meet the performance requirements of the application.}
\label{fig1}
\end{figure}

The execution of DNNs has two stages: training and inference. At the training stage, DNNs learn the rules to execute specific tasks from a corresponding dataset. During this process, millions of DNN parameters are adjusted, hence training is usually executed on cloud GPU(s). At the inference stage, the DNN is loaded with pre-trained parameters to execute the tasks, and the parameters are not changed. Inference can be executed on cloud GPU(s), where the data is sent from end-users to the cloud for processing and the inference result is returned. However, there is increasing interest in moving inference to be executed locally on mobile and embedded platforms due to a number of advantages. First, executing DNNs on the device can offer lower latency than executing them in the cloud. This lower latency leads to improved user experience, and/or the ability to meet the strict timing requirements of real-time applications such as self-driving cars \cite{sze2017efficient}. Second, mobile and embedded devices often collect and process personal data; keeping these data locally helps to mitigate potential privacy concerns \cite{alexa}. Finally, in areas with slow, intermittent or non-existent internet connectivity, timely cloud access can be difficult. Performing inference locally allows this issue to be mitigated.

It is attractive to be able to deploy DNN inference on a variety of platforms with distinct different computing resources, and to meet diverse application requirements (Fig \ref{fig1}). However, the high accuracy of DNNs comes at the cost of high computational requirements. DNNs are often too computationally intensive for resource-constrained platforms like mobile and embedded platforms\cite{NetAdapt, reform}, and this makes their efficient execution demanding and challenging. To efficiently execute DNNs on mobile and embedded platforms, a significant amount of recent work has focused on specialist hardware accelerator (also known as a Neural Processing Unit (NPU)) design \cite{chen2014diannao,chen2019eyeriss,han2016eie}. Design-time/offline approaches such as static model pruning\cite{NetAdapt,he2018amc} have been proposed to compress DNN models according to the application requirements and target hardware. However, while this can be applied offline to compress the model to approximately the `right size', managing the interface between available hardware resources (often numerous and heterogeneous in nature), software requirements, and user experience all of which are typically intractable at design-time. At runtime, since modern System on Chips (SoCs) typically execute a combination of different and dynamic workloads concurrently, it is challenging to consistently meet inference time/energy budgets because the local computing resources available to the DNN vary considerably (Fig \ref{fig2}). A variety of dynamic DNNs have been proposed to dynamically change the number of active filters to trade-off accuracy for time/energy reduction \cite{reform, tann2016runtime}. However, these approaches did not explore optimisation opportunities in hardware (e.g. Dynamic Voltage and Frequency Scaling (DVFS) and task mapping).

In this paper, we motivate the opportunities in online resource management for DNN inference on mobile and embedded platforms (Section \ref{motivation}), and explore how state-of-the-art approaches are enabling the dynamic performance scaling of DNNs that can be applied (Section \ref{review}). In particular, we use a case study (Section \ref{case_study}) to show how a Deep Neural Network (DNN) can be dynamically scalable to trade-off these various performance metrics. We then identify the opportunities for runtime resource management in this area, such that the performance trade-offs in both application and device can be explored and managed (Section \ref{opportunity}).

\section{Motivation for Online Resource Management}\label{motivation}
\begin{figure}
\centerline{\includegraphics[width=\columnwidth]{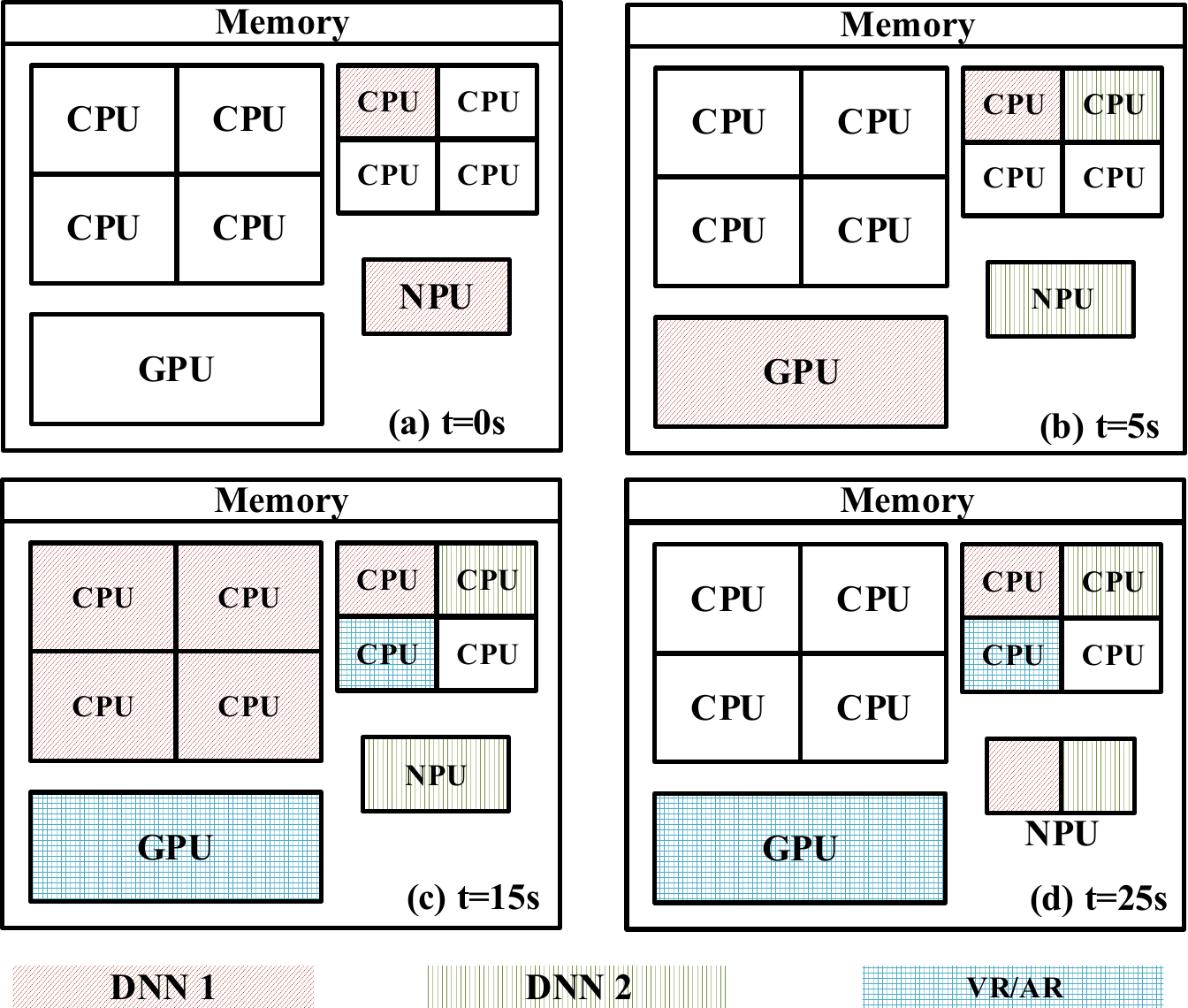}}
\caption{At runtime, the local computing resources available to DNNs may vary considerably due to multiple applications running concurrently.}
\label{fig2}
\end{figure}

Modern mobile and embedded SoCs typically contains multiple heterogeneous computing cores, such as CPUs, GPUs, DSPs, FPGAs, etc. In the last few years, the inclusion of NPU has become a trend to improve the efficiency of DNN inference execution. For example, the Huawei Kirin 990 5G \cite{kirin990} contains 8 CPU cores of three different types, a 16-core GPU and an NPU with 3 cores of two different types. The Apple A13 Bionic\cite{applea13} contains 6 CPU cores of two different types, a quad-core GPU and an NPU with 8 cores. In addition, dedicated matrix multipliers are attached to the CPU cluster for the acceleration of machine learning workloads.

It is attractive to be able to deploy DNN applications on a variety of different hardware platforms while consistently meeting pre-defined performance requirements. The performance of inference can be defined using platform-dependent (e.g. speed, energy) and platform-independent (accuracy, confidence) metrics. As shown in Table \ref{tab1}, when the same DNN model is deployed across different hardware platforms/cores, the execution time, energy and power consumption of inference vary considerably, but the accuracy remains the same. 

Two main challenges exist at design time and runtime. Different hardware platforms have vastly different computing resources and capabilities, and different applications have different performance requirements. Approaches are therefore needed to adapt DNN models to different hardware accordingly. This can be solved, at design-time (Fig \ref{fig1}), by `compressing' the DNN for target platform. For example, the same DNN might be used uncompressed on one platform with an NPU, while a compressed model (with offering lower accuracy but requiring fewer computations) is deployed on a different platform containing only CPU and GPU cores in order to meet the same execution time and energy consumption requirements.

\begin{table*}[htbp]
\begin{center}
\caption{Platform-dependent \& independent DNN performance metrics}
\begin{tabular}{|c|c|c|c|c|c|}
\cline{1-6} 
\multirow{2}{*}{\textbf{Platform}} & \multirow{2}{*}{\textbf{Computing cores}} & \multicolumn{3}{c|}{\textbf{Platform-dependent metrics}} & \textbf{Platform-independent metrics}\\
\cline{3-6} 
  &   & \textbf{Execution time (ms)} & \textbf{Power (mW)} & \textbf{Energy (mJ)} & \multicolumn{1}{c|}{\textbf{Top-1 Accuracy (\%)}} \\
\hline
\multirow{4}{*}{Jetson Nano}& GPU (614MHz) + A57 CPU (921MHz) & 7.4 & 1340 & 9.92 & \multirow{10}{*}{71.2}  \\
& GPU (921MHz) + A57 CPU (1.43GHz) & 4.93 & 2500 & 12.3 &  \\
& A57 CPU (921MHz) & 69.4 & 878 & 60.9 &   \\
& A57 CPU (1.43GHz) & 46.9 & 1490 & 69.9 &   \\
\cline{1-5} 
\multirow{6}{*}{Odroid XU3}& A15 CPU (200MHz) & 1020 & 326 & 320 &   \\
& A15 CPU (1GHz) & 204 &  846 & 173 &  \\
& A15 CPU (1.8GHz) & 117 &  2120 & 248 &  \\
& A7 CPU (200MHz) & 1780 &  72.4 & 129 &  \\
& A7 CPU (700MHz) & 504 &  141 & 71.4 &  \\
& A7 CPU (1.3GHz) & 280 &  329 & 92.1 & \\
\hline 
\end{tabular}
\label{tab1}
\end{center}
\end{table*}

While design-time approaches compress a model to a size that should be executable on the target platform, at runtime, the available computing resources may vary considerably due to multiple applications executing concurrently. Fig \ref{fig2} shows an example of how available computing resources may vary at runtime when different applications are executing:
\begin{itemize}
  \item \textbf{Single DNN [t=0s, Fig \ref{fig2}(a)]:} at the beginning, when there is only one DNN running on the platform, the NPU is used along with a CPU for pre-processing (e.g. image resizing). This may be the `compressed' model that was created for the platform at design time.
  \item \textbf{Two DNNs [t=5s, Fig \ref{fig2}(b)]:} a second DNN is deployed on the platform. It has higher requirements on the desired classification execution time, therefore it is executed on the NPU, and the first DNN is migrated to the GPU. Since GPU is typically slower than NPU for machine learning workloads, the first DNN is dynamically compressed which requires fewer computations by trading accuracy (see Section \ref{review}). This makes sure the performance requirements of two DNNs are both met.
  \item \textbf{Two DNNs and a VR/AR application [t=15s, Fig \ref{fig2}(c)]:} a VR/AR application is deployed onto the GPU, therefore the first DNN is migrated to the big CPU cluster and all four CPU cores are used due to the sheer volume of computations. Shortly after, the temperature of the SoC exceeds thermal limits. Therefore, the first DNN is dynamically compressed further and mapped onto a single core CPU in order to meet system thermal budgets.
  \item \textbf{Application performance requirement is changed [t=25s, Fig \ref{fig2}(d)]:} the accuracy requirement of the second DNN is reduced (e.g. by the user), therefore it can be dynamically compressed to a smaller model configuration and offers the spare NPU memory and computing capabilities to the first DNN. Two DNNs are dynamically scaled together to a model configuration where they are deployed on the same NPU.
\end{itemize}

Existing approaches for efficient execution of DNNs mainly focus on either hardware or software approaches. Although these works expose the optimisation opportunities on both sides, managing the interface between available hardware resources, software requirements, and user experience is not addressed, this identifies the opportunities for runtime resource management in this area.

\section{Approaches for Efficient Execution of DNNs}\label{review}
This section introduces state-of-the-art works on the efficient execution of DNNs from both the hardware and software perspectives. In particular, hardware accelerators for machine learning and static model pruning are widely used to address design time challenges, and dynamic model pruning approaches are used to let DNN consistently meet the performance requirements while running on dynamically available computing resources.

\subsection{Specialist DNN hardware accelerator}
A significant amount of work has been proposed from both academia (e.g. DianNao \cite{chen2014diannao}, EIE\cite{han2016eie} and Eyeriss v2 \cite{chen2019eyeriss}) and industry (e.g. Apple A13 Bionic\cite{applea13} and Huawei Kirin 990 5G\cite{kirin990}). 

In a typical DNN, the most computational intensive operation is matrix multiplication. Therefore, NPUs usually contain dedicated matrix multipliers, and since data movement dominates the energy consumption in DNN computation\cite{Chen:2016:ESA:3001136.3001177}, large on-chip memories are used to reduce off-chip accesses. It is attractive to be able to execute all DNNs on NPU, but not every hardware platform contains an NPU, and multiple DNNs executing concurrently will compete for the limited NPU resources at runtime. Mapping DNNs onto CPUs or GPUs may result in the violation of inference execution time and energy budgets. Therefore, DNN model pruning is also needed.

\begin{figure*}
\centerline{\includegraphics[width=\textwidth]{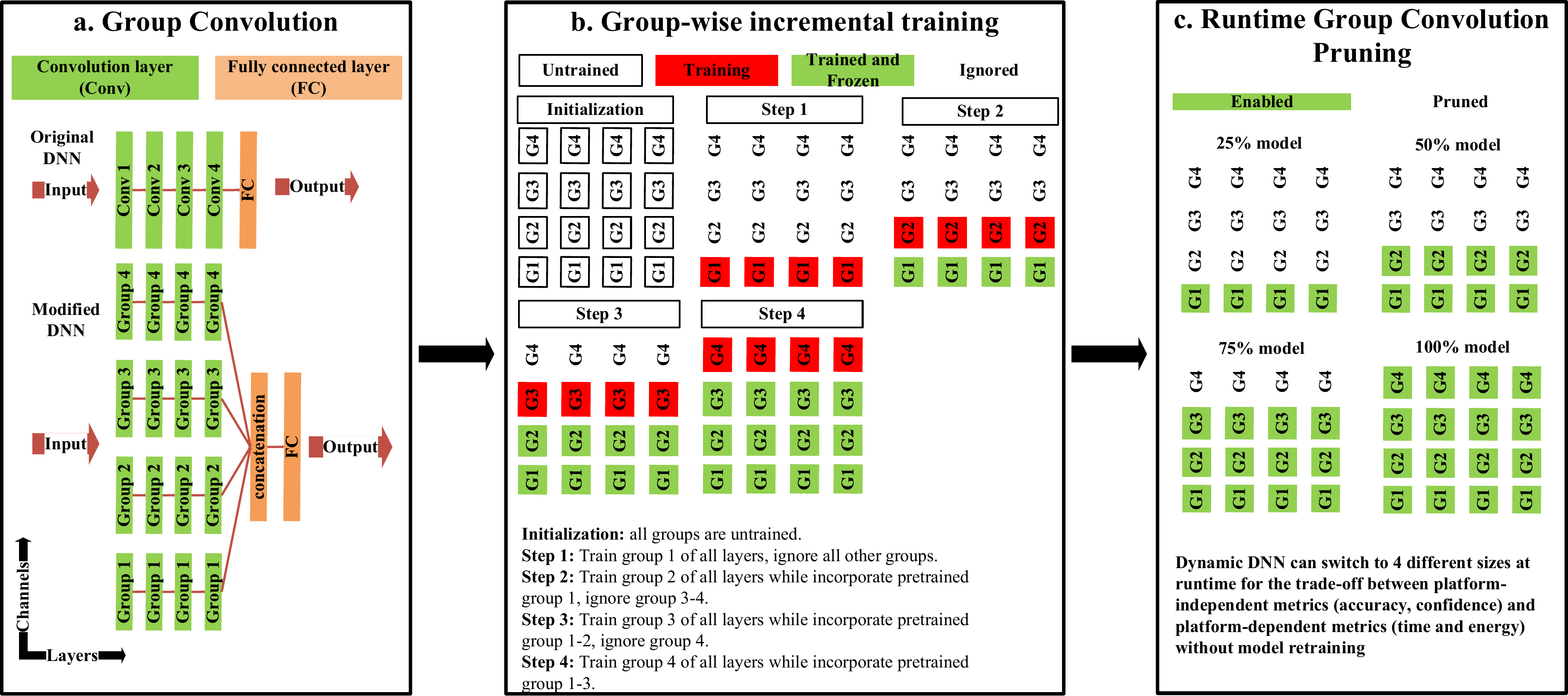}}
\caption{Dynamic DNN using incremental training and group convolution pruning. The (a) channels of the DNN convolution layer are divided into groups, which are then trained incrementally (b). At runtime, (c) later groups can be pruned for inference time/energy reduction or added back for accuracy recovery without model retraining.}
\label{fig3}
\end{figure*}

\subsection{Static Model Pruning}
Weight pruning is a static model compression approach to reduce the number of parameters in a DNN, Han \textit{et al.} \cite{pruning} proposed a magnitude-based algorithm, where parameters with small values are pruned. However, this approach leads to unstructured sparse filters which cannot be accelerated by most hardware\cite{Yu:2017:SCD:3079856.3080215}, except hardware accelerators that are designed specifically for sparse DNNs\cite{han2016eie,chen2019eyeriss}. Unlike weight pruning, filter pruning\cite{li2016pruning} prunes whole filters to compress the model. Although this approach has a lower model compression rate compared to weight pruning, it does not generate unstructured filters, and hence can gain actual acceleration on all hardware platforms. 

Yang \textit{et al.} \cite{NetAdapt} use filter pruning to compress DNNs to a meet pre-defined inference execution time budget on any target hardware platform. This approach offers a trade-off between accuracy and execution time. To achieve consistent performance across different platforms, the DNN is compressed more on platforms with less computing capabilities (i.e. sacrificing more accuracy for a reduction in execution time). For example, the full DNN model can be deployed on an NPU, yet the smaller compressed models can be deployed on GPUs and CPUs to meet the same time budget with less accuracy. This approach generates one DNN for a given performance budget at a pre-defined hardware setting (e.g. computing core and voltage/frequency level). 

This raises a significant problem, since modern SoCs typically execute a combination of different dynamic workloads concurrently, and hence the local resources available to the DNN vary considerably at runtime. The performance budgets cannot be met when the pre-defined hardware setting is unavailable at runtime. For example, the computing core may be unavailable because other applications are running on it, or available at a lower voltage/frequency due to other computing cores executing in the same voltage/frequency domain, or fixed power/thermal budgets. Multiple DNNs are needed to cover all hardware settings, which result in significant memory storage overhead. Furthermore, the switching activities of these DNNs at runtime may cause significant delay and energy consumption\cite{park2015big}.

\subsection{Dynamic Model Pruning}\label{dynamicDNN}
Static model pruning generates one DNN for a given performance budget and hardware setting combination. Multiple DNNs need to be generated to cover all hardware combinations (i.e. core, voltage and frequency) in an SoC. Dynamic Model Pruning (also known as dynamic DNNs) contains multiple DNN configurations in a single model. These configurations use a different number of filters within the same model, hence they are stored within a single model memory footprint and have different sizes, accuracies and computation requirements.

Dynamic DNNs \cite{lin2017runtime, reform, tann2016runtime, fang2018nestdnn} can be partially executed to consistently meet the performance budget (e.g. time, energy), while adapting to runtime resource varieties on the hardware. For example, smaller DNN configurations that are less accurate but require less computation are deployed when the computing capabilities of the hardware resources available at runtime are reduced (e.g. at a lower voltage/frequency). 

\section{Case Study: Exploring Performance Trade-offs with Dynamic DNNs} \label{case_study} 
Modern mobile and embedded SoCs are highly efficient because of the use of runtime resource management techniques, such as scheduling task mapping, DPM and DVFS. Although the dynamic DNNs presented above are scalable to trade-off various performance metrics, they can be combined with task mapping and DVFS to achieve a wider dynamic range of performance trade-off. To illustrate this, we proposed a dynamic DNN that can be scaled with task mapping and DVFS\cite{xun2019GCP}. 
The DNN is built using incremental training and group convolution pruning, and is shown in Fig \ref{fig3}. The channels of the DNN's convolution layers are divided into groups (Fig \ref{fig3} (a)), which are then trained incrementally (Fig \ref{fig3} (b)). At runtime, later groups can be pruned for inference time/energy reduction, or added back for accuracy recovery when more computing resources become available. This is all possible at runtime without model retraining (Fig \ref{fig3} (c)). 

\begin{figure*}
\centerline{\includegraphics[width=\textwidth]{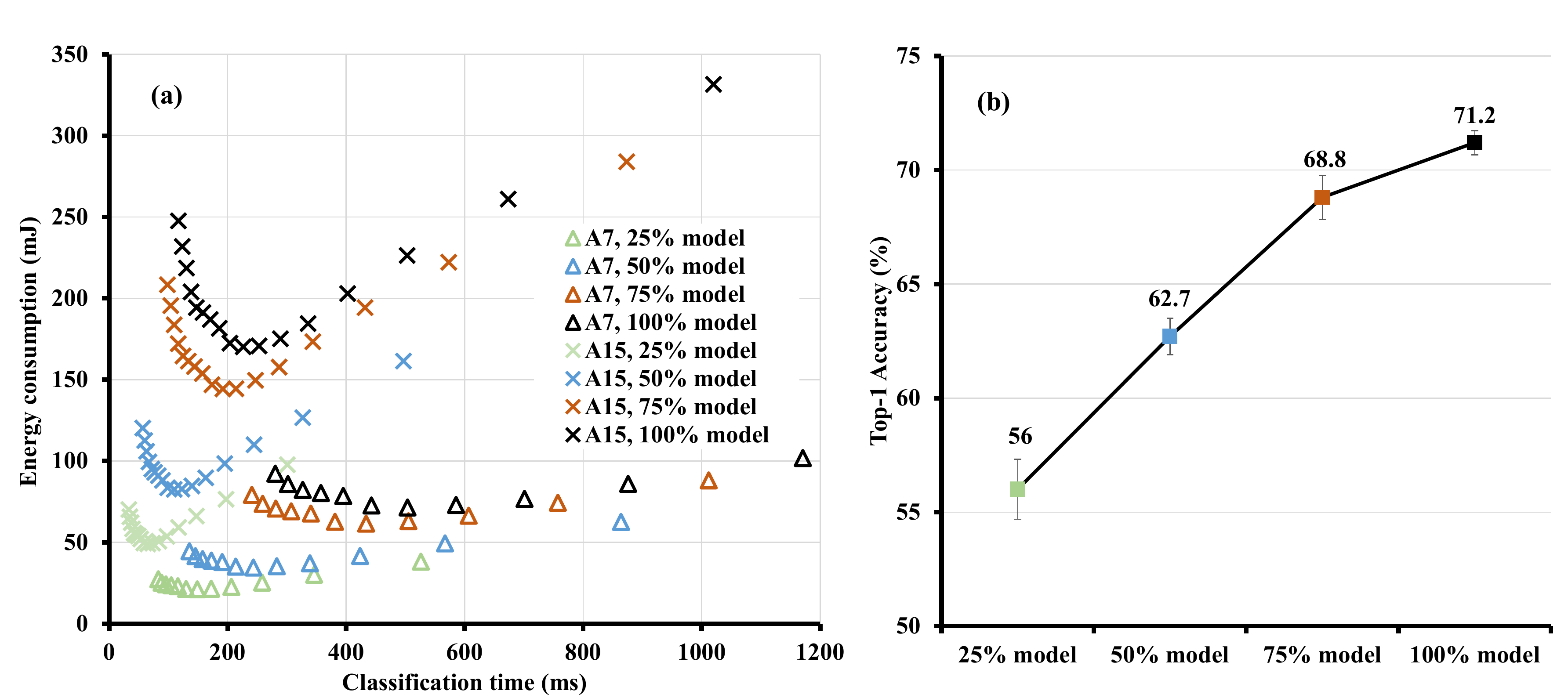}}
\caption{(a) E, t operating points sapce. A Dynamic DNN (different colours show different levels of `compression') is combined with task mapping (different symbols) and DVFS (different points) on the Odroid XU3 platform. (b) Top-1 image classification accuracy on 10,000 CIFAR10 validation images. The error bar shows the variance over 10 image classes of CIFAR10.}
\label{fig4}
\end{figure*}

We use a four-increment design to obtain four different DNN configurations, and refer them as the 25\%, 50\%, 75\% and 100\% models. The impact on classification accuracy is shown in Fig \ref{fig4}(b). The 25\% model uses only one group of DNN parameters; therefore it is the least accurate model but requires the minimum computation. The 100\% model is the full model; therefore it is the most accurate and computationally expensive model. The design is validated through empirical measurements on the Odroid XU3 heterogeneous multi-core platform. The DNN is mapped on both Arm A15 and A7 CPU cores, and under 17 and 12 different frequency levels respectively. The results are shown in Fig \ref{fig4}(a). Task mapping, DVFS and the dynamic DNN can be seen as three adjustable knobs which can be adjusted to meet dynamic E, P, t and accuracy budgets/targets at runtime. For example, in Fig \ref{fig4}, for a budget of 400 ms and 100 mJ, a 100\% model on the A7 CPU at 900 MHz could offer the highest accuracy and lowest energy consumption. If the budgets change to 200 ms and 150 mJ, then a 75\% model on the A15 CPU at 1 GHz becomes the new optimal configuration. In the case of multiple applications executing concurrently, the frequency setting may be sub-optimal due to other applications in the same frequency domain are using that frequency level.

\section{Online Resource Management of Machine Learning}  \label{opportunity}

The above approaches offer a number of dynamically selectable operating point in the E, P, t, accuracy space. However, to meet the opportunities of Fig \ref{fig2}, these approaches need to be integrated with online resource management.

A variety of online resource management approaches have been proposed, such as DVFS \cite{das2014reinforcement}, task mapping \cite {reddy2017inter} and power gating \cite{rahmani2016reliability}. These approaches optimise hardware behaviour to satisfy constraints (e.g. temperature, power); the performance requirements and optimisation opportunities in the application are traditionally not addressed. 

Runtime management (RTM) can be enhanced by using `knobs' and `monitors' of the application and device, which provide interfaces to convey information between RTM, applications and devices. A variety of works have been proposed \cite{donyanavard2019sosa, moazzemi2019hessle, gadioli2015application, fleming2014heterogeneous}, which focus on the opportunities in either applications or devices. It is necessary to explore the opportunities on both sides at the same time to address the sheer volume of computation in DNN applications. Bragg \textit{et al.} \cite{bragg2018application} proposed the PRiME framework, that abstracts the system into three layers: application, device and RTM; control between them operates through knobs and monitors in the application and hardware devices. In particular, knobs are adjustable parameters in the application (e.g. execution iteration, data precision) and device (e.g. voltage/frequency, core), and monitors offer performance information about the application (e.g. accuracy, frame rate) and device (e.g. power, temperature). 



\begin{figure}
\centerline{\includegraphics[width=\columnwidth]{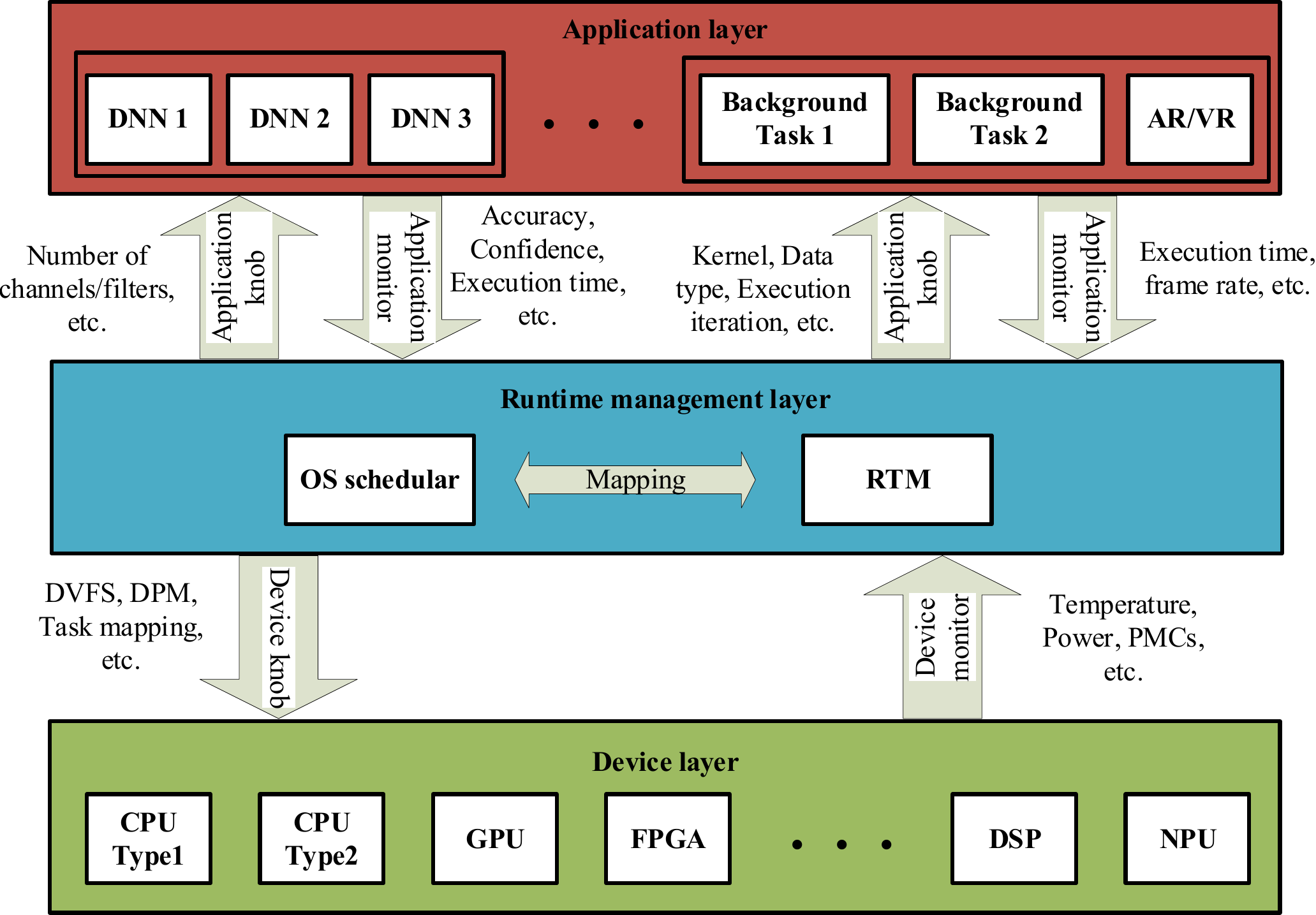}}
\caption{Apply state-of-the-art runtime resource management for embedded machine learning. The DNN performance trade-off opportunities in the hardware and software are managed through knobs and monitors controlled by RTM.}
\label{fig5}
\end{figure}

Fig \ref{fig5} shows how dynamic DNN, task mapping and DVFS can be combined together alongside an RTM framework such as PRiME. The opportunities for DNN performance trade-off in both hardware and software are managed through knobs and monitors that are controlled by RTM. For example, dynamic DNN (application knob) can be scaled with DVFS and task mapping (device knobs) to meet the DNN performance requirements (application monitors), and meet any hardware temperature and power constraints (device monitors). In the case of Fig \ref{fig2} (d), the RTM received new accuracy requirements for the DNNs through the application monitor, and hence RTM changed the size of the DNNs using the application knob, mapping them onto the NPU using the device knob. In addition, system physical limits like temperature and power are monitored using the device monitors, as DVFS could be then applied in order to meet these limits.

\section{Conclusions}
This paper has presented the challenges of deploying DNNs on mobile and embedded platforms, at both design time and runtime. At design time, DNNs are deployed on a variety of hardware platforms with vastly different computing resources to meet different application requirements. However, at runtime, it is very challenging to consistently meet performance requirements, since the availability of the local computing resources vary considerably due to other applications executing concurrently. We showed how approaches enabling the dynamic performance scaling of DNNs can be applied to address these challenges, and proposed how online resource management approaches can be applied to manage and optimise machine learning workloads alongside other applications at runtime. Execution of DNN inference on mobile and embedded platforms is clearly important, but also challenging, and we believe that further research is necessitated in runtime resource allocation and adaptation to optimise this.

\section{Acknowledgement}

This work was supported in part by the Engineering and Physical Sciences Research Council (EPSRC) under Grant EP/S030069/1. Experimental data can be found at DOI: 10.5258/SOTON/D1154


\begin{thebibliography}{10}
\providecommand{\url}[1]{#1}
\csname url@samestyle\endcsname
\providecommand{\newblock}{\relax}
\providecommand{\bibinfo}[2]{#2}
\providecommand{\BIBentrySTDinterwordspacing}{\spaceskip=0pt\relax}
\providecommand{\BIBentryALTinterwordstretchfactor}{4}
\providecommand{\BIBentryALTinterwordspacing}{\spaceskip=\fontdimen2\font plus
\BIBentryALTinterwordstretchfactor\fontdimen3\font minus
  \fontdimen4\font\relax}
\providecommand{\BIBforeignlanguage}[2]{{%
\expandafter\ifx\csname l@#1\endcsname\relax
\typeout{** WARNING: IEEEtran.bst: No hyphenation pattern has been}%
\typeout{** loaded for the language `#1'. Using the pattern for}%
\typeout{** the default language instead.}%
\else
\language=\csname l@#1\endcsname
\fi
#2}}
\providecommand{\BIBdecl}{\relax}
\BIBdecl

\bibitem{deeplearning}
Y.~LeCun, Y.~Bengio, and G.~Hinton, ``Deep learning,'' \emph{Nature}, vol. 521,
  no. 7553, p. 436, 2015.

\bibitem{alexnet}
A.~Krizhevsky, I.~Sutskever, and G.~E. Hinton, ``Imagenet classification with
  deep convolutional neural networks,'' in \emph{Advances in Neural Information
  Processing Systems (NeurIPS)}, 2012, pp. 1097--1105.

\bibitem{yolo}
J.~Redmon, S.~Divvala, R.~Girshick, and A.~Farhadi, ``You only look once:
  Unified, real-time object detection,'' in \emph{Conference on Computer Vision
  and Pattern Recognition (CVPR)}, 2016, pp. 779--788.

\bibitem{vaswani2017attention}
A.~Vaswani, N.~Shazeer, N.~Parmar, J.~Uszkoreit, L.~Jones, A.~N. Gomez,
  {\L}.~Kaiser, and I.~Polosukhin, ``Attention is all you need,'' in
  \emph{Advances in Neural Information Processing Systems (NeurIPS)}, 2017, pp.
  5998--6008.

\bibitem{NetAdapt}
T.-J. Yang \emph{et~al.}, ``Netadapt: Platform-aware neural network adaptation
  for mobile applications,'' in \emph{European Conference on Computer Vision
  (ECCV)}, 2018, pp. 285--300.

\bibitem{he2018amc}
Y.~He, J.~Lin, Z.~Liu, H.~Wang, L.-J. Li, and S.~Han, ``Amc: Automl for model
  compression and acceleration on mobile devices,'' in \emph{European
  Conference on Computer Vision (ECCV)}, 2018, pp. 784--800.

\bibitem{sze2017efficient}
V.~Sze, Y.-H. Chen, T.-J. Yang, and J.~S. Emer, ``Efficient processing of deep
  neural networks: A tutorial and survey,'' \emph{Proceedings of the IEEE},
  vol. 105, no.~12, pp. 2295--2329, 2017.

\bibitem{alexa}
D.~Lynskey, ``{Alexa, are you invading my privacy?}'' [Online]. Available:
  \url{https://www.theguardian.com/technology/2019/oct/09/alexa-are-you-invading-my-privacy-the-dark-side-of-our-voice-assistants
  }.

\bibitem{reform}
Z.~Xu, F.~Yu, C.~Liu, and X.~Chen, ``Reform: Static and dynamic resource-aware
  dnn reconfiguration framework for mobile device,'' in \emph{Design Automation
  Conference (DAC)}, 2019, p. 183.

\bibitem{chen2014diannao}
T.~Chen, Z.~Du, N.~Sun, J.~Wang, C.~Wu, Y.~Chen, and O.~Temam, ``Diannao: A
  small-footprint high-throughput accelerator for ubiquitous
  machine-learning,'' in \emph{ASPLOS}, 2014, pp. 269--284.

\bibitem{chen2019eyeriss}
Y.-H. Chen, T.-J. Yang, J.~Emer, and V.~Sze, ``Eyeriss v2: A flexible
  accelerator for emerging deep neural networks on mobile devices,'' \emph{IEEE
  Journal on Emerging and Selected Topics in Circuits and Systems}, 2019.

\bibitem{han2016eie}
S.~Han, X.~Liu, H.~Mao, J.~Pu, A.~Pedram, M.~A. Horowitz, and W.~J. Dally,
  ``{EIE: efficient inference engine on compressed deep neural network},'' in
  \emph{International Symposium on Computer Architecture (ISCA)}, 2016, pp.
  243--254.

\bibitem{tann2016runtime}
H.~Tann, S.~Hashemi, R.~Bahar, and S.~Reda, ``Runtime configurable deep neural
  networks for energy-accuracy trade-off,'' in \emph{International Conference
  on Hardware/Software Codesign and System Synthesis}.\hskip 1em plus 0.5em
  minus 0.4em\relax ACM, 2016, p.~34.

\bibitem{kirin990}
``{Huawei kirin 990 series},'' [Online]. Available:
  \url{https://consumer.huawei.com/en/campaign/kirin-990-series/}.

\bibitem{applea13}
J.~Cross, ``{Inside Apple’s A13 Bionic system-on-chip},'' [Online].
  Available:
  \url{https://www.macworld.com/article/3442716/inside-apples-a13-bionic-system-on-chip.html}.

\bibitem{Chen:2016:ESA:3001136.3001177}
Y.-H. Chen, J.~Emer, and V.~Sze, ``Eyeriss: A spatial architecture for
  energy-efficient dataflow for convolutional neural networks,'' in
  \emph{International Symposium on Computer Architecture (ISCA)}, 2016, pp.
  367--379.

\bibitem{pruning}
S.~Han, J.~Pool, J.~Tran, and W.~Dally, ``Learning both weights and connections
  for efficient neural network,'' in \emph{Advances in Neural Information
  Processing Systems (NeurIPS)}, 2015, pp. 1135--1143.

\bibitem{Yu:2017:SCD:3079856.3080215}
J.~Yu, A.~Lukefahr, D.~Palframan, G.~Dasika, R.~Das, and S.~Mahlke, ``Scalpel:
  Customizing dnn pruning to the underlying hardware parallelism,'' in
  \emph{International Symposium on Computer Architecture (ISCA)}, 2017, pp.
  548--560.

\bibitem{li2016pruning}
H.~Li, A.~Kadav, I.~Durdanovic, H.~Samet, and H.~P. Graf, ``Pruning filters for
  efficient convnets,'' \emph{arXiv preprint arXiv:1608.08710}, 2016.

\bibitem{park2015big}
E.~Park, D.~Kim, S.~Kim, Y.-D. Kim, G.~Kim, S.~Yoon, and S.~Yoo, ``Big/little
  deep neural network for ultra low power inference,'' in \emph{International
  Conference on Hardware/Software Codesign and System Synthesis}.\hskip 1em
  plus 0.5em minus 0.4em\relax IEEE Press, 2015, pp. 124--132.

\bibitem{lin2017runtime}
J.~Lin, Y.~Rao, J.~Lu, and J.~Zhou, ``Runtime neural pruning,'' in
  \emph{Advances in Neural Information Processing Systems (NeurIPS)}, 2017, pp.
  2181--2191.

\bibitem{fang2018nestdnn}
B.~Fang, X.~Zeng, and M.~Zhang, ``Nestdnn: Resource-aware multi-tenant
  on-device deep learning for continuous mobile vision,'' in
  \emph{International Conference on Mobile Computing and Networking}.\hskip 1em
  plus 0.5em minus 0.4em\relax ACM, 2018, pp. 115--127.

\bibitem{xun2019GCP}
L.~Xun, L.~Tran-Thanh, B.~M. Al-Hashimi, and G.~V. Merrett, ``Incremental
  training and group convolution pruning for runtime dnn performance scaling on
  heterogeneous embedded platforms,'' in \emph{Workshop on Machine Learning for
  CAD (MLCAD)}, 2019.

\bibitem{das2014reinforcement}
A.~Das, R.~A. Shafik, G.~V. Merrett, B.~M. Al-Hashimi, A.~Kumar, and
  B.~Veeravalli, ``Reinforcement learning-based inter-and intra-application
  thermal optimization for lifetime improvement of multicore systems,'' in
  \emph{Design Automation Conference (DAC)}, 2014, pp. 1--6.

\bibitem{reddy2017inter}
B.~K. Reddy, A.~K. Singh, D.~Biswas, G.~V. Merrett, and B.~M. Al-Hashimi,
  ``Inter-cluster thread-to-core mapping and dvfs on heterogeneous
  multi-cores,'' \emph{Transactions on Multi-Scale Computing Systems}, vol.~4,
  no.~3, pp. 369--382, 2017.

\bibitem{rahmani2016reliability}
A.~M. Rahmani, M.-H. Haghbayan, A.~Miele, P.~Liljeberg, A.~Jantsch, and
  H.~Tenhunen, ``Reliability-aware runtime power management for many-core
  systems in the dark silicon era,'' \emph{IEEE Transactions on Very Large
  Scale Integration (VLSI) Systems}, vol.~25, no.~2, pp. 427--440, 2016.

\bibitem{donyanavard2019sosa}
B.~Donyanavard, T.~M{\"u}ck, A.~M. Rahmani, N.~Dutt, A.~Sadighi, F.~Maurer, and
  A.~Herkersdorf, ``Sosa: Self-optimizing learning with self-adaptive control
  for hierarchical system-on-chip management,'' in \emph{International
  Symposium on Microarchitecture (MICRO)}.\hskip 1em plus 0.5em minus
  0.4em\relax ACM, 2019, pp. 685--698.

\bibitem{moazzemi2019hessle}
K.~Moazzemi, B.~Maity, S.~Yi, A.~M. Rahmani, and N.~Dutt, ``Hessle-free:
  Heterogeneous systems leveraging fuzzy control for runtime resource
  management,'' \emph{ACM Transactions on Embedded Computing Systems (TECS)},
  vol.~18, no.~5s, p.~74, 2019.

\bibitem{gadioli2015application}
D.~Gadioli, G.~Palermo, and C.~Silvano, ``Application autotuning to support
  runtime adaptivity in multicore architectures,'' in \emph{International
  Conference on Embedded Computer Systems: Architectures, Modeling, and
  Simulation (SAMOS)}.\hskip 1em plus 0.5em minus 0.4em\relax IEEE, 2015, pp.
  173--180.

\bibitem{fleming2014heterogeneous}
S.~T. Fleming and D.~B. Thomas, ``Heterogeneous heartbeats: A framework for
  dynamic management of autonomous socs,'' in \emph{International Conference on
  Field Programmable Logic and Applications (FPL)}.\hskip 1em plus 0.5em minus
  0.4em\relax IEEE, 2014, pp. 1--6.

\bibitem{bragg2018application}
G.~M. Bragg, C.~R. Leech, D.~Balsamo, J.~J. Davis, E.~Weber~Wachter,
  G.~Merrett, G.~A. Constantinides, and B.~Al-Hashimi, ``An application-and
  platform-agnostic control and monitoring framework for multicore systems,''
  in \emph{International Conference on Pervasive and Embedded Computing (PEC)},
  2018.

\end{thebibliography}


\end{document}